\documentclass[a4paper,twoside]{article}

\usepackage{epsfig}
\usepackage{subfigure}
\usepackage{calc}
\usepackage{amssymb}
\usepackage{amstext}
\usepackage{amsmath}
\usepackage{amsthm}
\usepackage{multicol}
\usepackage{pslatex}
\usepackage{apalike}
\usepackage[small]{caption}
\usepackage[linesnumbered]{algorithm2e}
\usepackage{SCITEPRESS}
\makeatletter
\renewcommand{\@algocf@capt@plain}{above}
\makeatother

\begin{document}

\title{Robust method of vote aggregation and proposition verification for invariant local features}

\author{\authorname{Grzegorz Kurzejamski, Jacek Zawistowski, Grzegorz Sarwas}
\affiliation{Lingaro Sp. z o.o.\\
Pu\l awska 99a, 02-595 Warsaw, Poland\\}
\email{ \{grzegorz.kurzejamski, jzawisto, grzegorz.sarwas\}@gmail.com}
}

\keywords{Computer Vision, Image Analysis, Multiple Object Detection, Object Localization, Pattern Matching}

\abstract{This paper presents a method for analysis of the vote space created from the local features extraction process in a multi-detection system. The method is opposed to the classic clustering approach and gives a high level of control over the clusters composition for further verification steps. Proposed method comprises of the graphical vote space presentation, the proposition generation, the two-pass iterative vote aggregation and the cascade filters for verification of the propositions. Cascade filters contain all of the minor algorithms needed for effective object detection verification. The new approach does not have the drawbacks of the classic clustering approaches and gives a substantial control over process of detection. Method exhibits an exceptionally high detection rate in conjunction with a low false detection chance in comparison to alternative methods.}

\onecolumn \maketitle \normalsize \vfill

\section{\uppercase{Introduction}}

\noindent Object detection based on local features is well known in the computer vision field. Many different researches brought about different features and methods of scene analysis in search of a particular object. Recently developed feature points are proven to be well suited for a specific object detection, rather than a generalised object's class identification. As many applications of local features has been evaluated, the ability to describe selective elements of rich graphics is the main purpose of invariant local features in many fields. Amongst the most popular local features are \emph{e.g.}\ SIFT, SURF, BRISK, FREAK, MSER. These are commonly described as feature points or feature regions. They are easy to manipulate and to match. Use of such local features gives not only the ability to describe the object in many ways, but also to achieve invariance for basic object and image transformation, as skew, rotation, blur, noise. Invariant local features used in conjunction with invariant characteristic region detectors, such as Harris-Affine or SIFT detector, provide data for thorough scene-object analysis, leading to detection of an object in the scene. Giving an appropriate set of features one can determine the exact position of the object in the scene with a specified scale, rotation and even minor, linear deformations. 

The straightforward approach to a detection task is to identify local features in the scene and the pattern and match them against themselves, using an appropriate metric. Common evaluations of such systems use brute KNN classification or its derivatives as FLANN KNN search or BBF search. These give good approximation of, theoretically ideal, results with substantially lower computational cost. Some of the applications make use of a LSH hashing but this approach does not provide practical distance data needed for further analysis. Matching local features provides set of correspondences, that can be filtered later. Under the assumption, that the scene contains one or no instance of the object, correspondence can be put into the object-related or noise-related class. To distinguish which class a particular correspondence belongs to, few methods have been developed. Frequently used method for such purpose is RANSAC, giving very good results, even for a high level of noise-related correspondences. After the classification of correspondences the model can be assumed and a homography can be calculated. Well designed parameters and filters can lead to high detection rate and low false detection chance. RANSAC and quantitative analysis of correspondence data become inefficient, when a contribution of noise-related correspondences in whole correspondence group grows. Because of that such approach is not sufficient for scenes, where the objects occupy small share of the image. Methods mentioned above will not work well with multiple objects present in the image as well.

Systems for multi-detection purposes incorporate divide and conquer approach. Each correspondence can be assigned to one of N+1 classes, where N is the number of objects in the scene. There is no known, straightforward method of assigning correspondences. Most applications use correspondences as votes in a multidimensional space. Vote space can be clustered with common clustering algorithms. Each cluster can be processed with a single-object detection algorithm. Clustering approaches can be divided into two groups. The first one consists of sparse clustering, where each cluster should contain all the needed vote data of a specific class. Such methods show low detection rate because of a far from ideal clustering process and a high level of parametrization. Second group is dense clustering, where clusters may contain only small portion of a particular correspondence class. Its analysis leads to creating or supporting hypothesis of the object's occurrence in the scene. The flagships in that matter are Hough-like methods. Such approaches show high detection rate but high false positive rate as well. There are some works that try to segment the scene with known context, as shown in the work of Iwanowski \emph{et al.}~\cite{Polki01}.

In our tests both clustering approaches lacked the ability to attain a very high detection rate with a very low false positive rate at the same time. For our test cases a processing power is not a limitation and the images are of a very high quality. Our test data contains from zero to up to 100 objects per image and presents different environmental conditions. Most of the stat-of-the-art publications do not test detection capabilities for such complex tasks. We found that current approaches cannot maintain good detection rate to false positive rate ratio on satisfactory level in many real life applications. 

This paper presents the method of vote space analysis, a part of invention shown in~\cite{Pse01}. The method can be adjusted to a vast variety of object detection purposes, where the effectiveness and a low false positive rate is crucial. The method has been developed to work well with huge amount of feature data, extracted from high quality images. Most of the algorithms used in the new approach come with a logical justification. The new method uses the iterative vote aggregation, starting from proposition's positions. Propositions are generated from a graphical vote space analysis. Aggregated data undergoes analysis and filtration. Whole process has a two-pass model, that makes the method robust to some specific object positioning in the scene. Cascade of a specially selected set of filter algorithms has been utilized to reject most of the false positive detections.

\section{\uppercase{Related work}}

\noindent Local features in the image can be tracked a long way in the literature. We present state-of-the-art feature points extracting and describing methods, that can be used in our method, and similar frameworks for multi-object detection purposes developed through the last years.

\subsection{Feature points}
Our method should be used with conjunction with scale-invariant and rotation-invariant features for the best results. Usage of local features lacking any of this characteristics may come with a need for rejection of some parts of our method, but can be implemented nevertheless.

The best known feature points, up to this point, are SIFT points developed by Lowe~~\cite{Sift01}, which became a model for various local features benchmarks. The closest alternative to SIFT is SURF~\cite{Surf01}, that comes with a lower dimensionality and, in the result, a higher computing efficiency. There are also known attempts to incorporate additional enhancements into SIFT and SURF as PCA-SIFT~\cite{Pca-sift01} or Affine-SIFT~\cite{Asift01}. SIFT and SURF and its derivatives are computationally demanding during matching process. In last years there has been big development in feature points based on binary test pairs, that can be matched and described in a very fast manner. The flagships of this approach are BRIEF~\cite{Brief01}, ORB~\cite{Orb01}, BRISK~\cite{Brisk01} and FREAK~\cite{Freak01} features. Most of the cited algorithms can be used to create dense and highly discriminative voting space, which holds substantial object correspondence data needed to accomplish many of the real-world detection tasks.

\subsection{Frameworks}

There are few approaches to conduct multi-object multi-detection, meaning detecting multiple different objects on the scene, where any object can be visible in multiple places. Viola and Jones~\cite{Haar01} developed cascade of boosted features, that can efficiently detect multiple instances of the same object in a one pass of the detection process. The method needs a time consuming, learning process with thousands of images. Method has been mostly tested on general objects, as people, cars, faces. Most straightforward method for multi-detection is using all of the sliding windows as used, for example, in Sarwas' and Skoneczny's work~\cite{VarFil01}. Most of them are unfortunately computationally expensive. High effectiveness can be achieved with Histogram of Oriented Gradients~\cite{Hog01} and Deformable Part Models~\cite{Dpm01}. The biggest drawbacks for our application is that Deformable Part Models needs learning stage and Histogram of Oriented Gradients is not rotation invariant. Blaschko and Lampert in~\cite{BlaschkoL08} uses SVM to enhance the sliding window process. Efficient subwindows search has been used in~\cite{LampertBH08}. In addition, branch-and-bound approaches, as in~\cite{Yeh09}, are promising for multi-detection purposes with conjunction with Bag-of-words descriptors. Lowe~\cite{Sift02} proposed generalized Hough Transform for clustering vote space with SIFT correspondence data. Authors of \cite{AzadAD09a} created a 4D voting space and used combination of Hough, RANSAC and Least Squares Homography Estimation in order to detect and accept potential object instances. Zickler in \emph{et al.}~\cite{ZicklerE07} used angle differences criterion in addition to RANSAC mechanisms and vote number threshold. Zickler \emph{et al.}\ in \cite{ZicklerV06} used custom probabilistic model in addition to Hough algorithm.

In our system's application we could use only one generic pattern image per object so we rejected most of the learning-based global descriptors. 

\section{\uppercase{Algorithm}}

The algorithm presented by authors is built upon two mechanisms: the vote spaces creation and a vote aggregation for each of the vote spaces created. The vote space is created for each pattern. Its adjacency data is projected onto the (X, Y) plane, creating vote images (one for each vote space). The vote images are analysed in search for object's position propositions. This mechanism is shown in the Algorithm \ref{alg:prop_creation}. The aggregation process is performed for each vote space and for its each proposition, starting from the proposition with the highest adjacency value. Aggregation consists of two passes with slightly different vote gathering approaches. The first pass is needed to estimate the detected object's area in the scene, so the second aggregation pass would gather only votes considered to be from that particular object's instance. The structure of each pass is presented in Algorithm \ref{alg:aggregation}.

\begin{algorithm}[]
 \KwData{Original Patterns (OPT), Scene Image (SCN)}
 \KwResult{ Vote Data and propositions for object's centres for each pattern. }
 
 Feature points extraction on OPT and SCN\;
 \ForEach{pattern in OPT}
 {
 Find correspondences (COR) between pattern and SCN feature points\;
 \ForEach{correspondence in COR}
 {
 Reject if has low distance value\;\label{line:distance_filter}
 Reject if has high hue difference value\;\label{line:color_filter}
 Calculate adjacency value\;\label{line:adjacency_calc}
 }
 Creation of vote space (VS) from COR\;\label{line:vs_from_cor}
 Creation of vote image (VI) from VS\;\label{line:vi_from_vs}
 Search for propositions (PR) in VI\;
 Sort PR list\;\label{line:prop_sorting}
 }
 
 \caption{Vote Data, Vote Image and propositions creation}
 \label{alg:prop_creation}
\end{algorithm}

\begin{algorithm}[]
 \KwData{VI, VS, PR}
 \KwResult{ Occurrences (OCR) in the SCN for a particular pattern}
 
 \ForEach{Proposition in PR}
 {
 Gather all votes in local area from VS\;\label{line:pass1_1}
 Unique filtering for gathered votes (V)\;\label{line:pass1_2}
 Cascade filtering for V\;\label{line:pass1_3}
 \eIf{not rejected by cascade filtering}
 {
 Estimate object's area\;\label{line:pass1_4}
 Gather all votes with a Flood Fill algorithm\;\label{line:pass2_1}
 Unique filtering for new V\;\label{line:pass2_2}
 Cascade filtering for new V\;\label{line:pass2_3}
 \eIf{not rejected by cascade filtering}
 {
 Calculate object's area\;\label{line:pass2_4}
 Create occurrence entry in OCR\;
 Erase all vote data in occurrence's area in VS and VI\;
 }
 {reject proposition}
 }{reject proposition}
 
 }
 
 \caption{Vote aggregation and detection acceptance}
 \label{alg:aggregation}
\end{algorithm}

\subsection{Vote image creation}

\noindent First part of our method is the vote space and vote image creation (lines \ref{line:vs_from_cor} and \ref{line:vi_from_vs} of Algorithm \ref{alg:prop_creation}). Vote space consists of multiple dimensions: X, Y, Scale, Rotation and Distance. Each vote contains specific X and Y position of the center of the object. The Distance may be the result of using specific metric for particular feature points. For SIFT the standard procedure is to use L2 distance for its feature vector, which contains gradient data in the area around the characteristic point. One may use some additional information in distance calculation, as color difference. Someone can use ranking method as the LSH hashing instead of the L2 metric as well.

Vote image is the projection of the adjacency data available in vote space onto X and Y dimensions. Vote image has one intensity channel, created by normalizing adjacency sum cue. Another approach would be to use the distance value instead of adjacency as the main cue. We found L2 metric, as well as many other distance-based approaches, as insufficient. 

Votes in the vote spaces are built upon filtered correspondence sets. The distance threshold used in line \ref{line:distance_filter} of Algorithm \ref{alg:prop_creation} was calculated as:

\begin{equation}
\mbox{\textit{thr}} = \frac{MIN(V) + MAX(V)}{2},
\end{equation}where V is the votes group and the MIN and MAX operators return the value of a vote with minimal and maximal distance value from the group. The rejection function D is presented in equation \ref{eq:D_v}.

\begin{equation}
 D(v) = \left\{
  \begin{array}{l l}
    accept, \quad {dist}(v) \leq thr\\
    reject, \quad {dist}(v) > thr
  \end{array} \right..
  \label{eq:D_v}
\end{equation}

We transform distance value into normalized adjacency value in range from 0 to 1 (line \ref{line:adjacency_calc} of Alg. \ref{alg:prop_creation}). 1 indicates perfect match. 0 indicates near to rejection difference between feature points. We transformed distance values into adjacency (\emph{adj}) values with a specific function:

\begin{equation}
 \mbox{\textit{adj}}(v) = 1 - \left(
 \frac{dist(v)}{thr}
  \right)^2.
\end{equation}

\begin{figure*}[!htb]
\begin{minipage}[!htb]{.3\linewidth}
  \centering
  \centerline{\includegraphics[width=5cm]{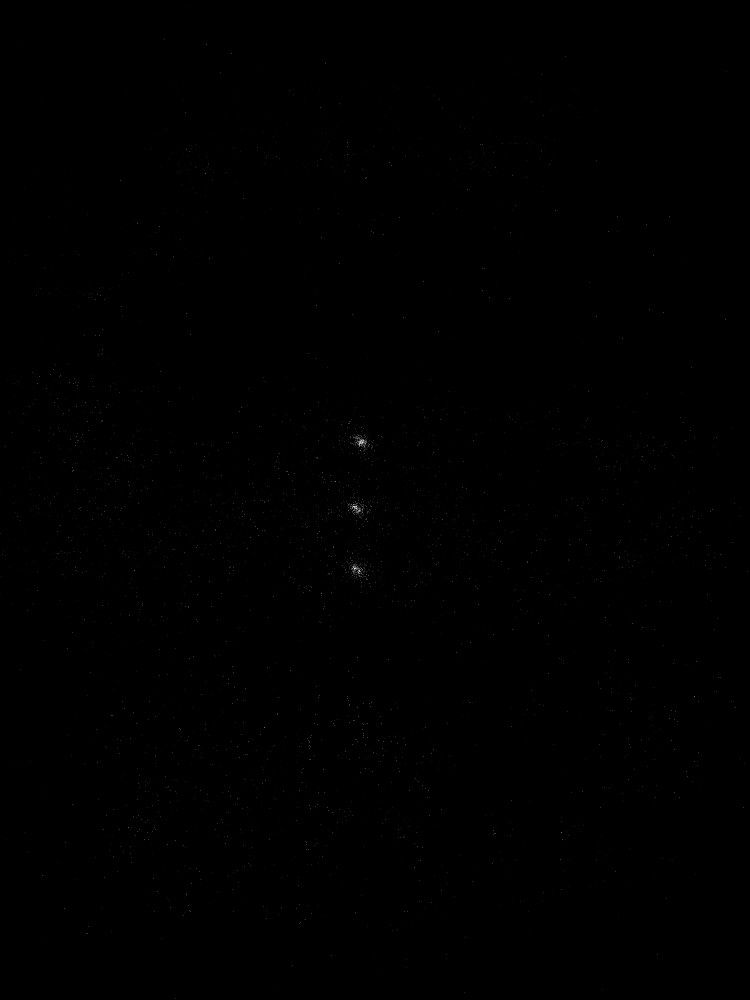}}
  \centerline{(a) Vote image}\medskip
  \label{fig:votemap_obj1}
\end{minipage}
\hfill
\begin{minipage}[!htb]{.3\linewidth}
  \centering
  \centerline{\includegraphics[width=5cm]{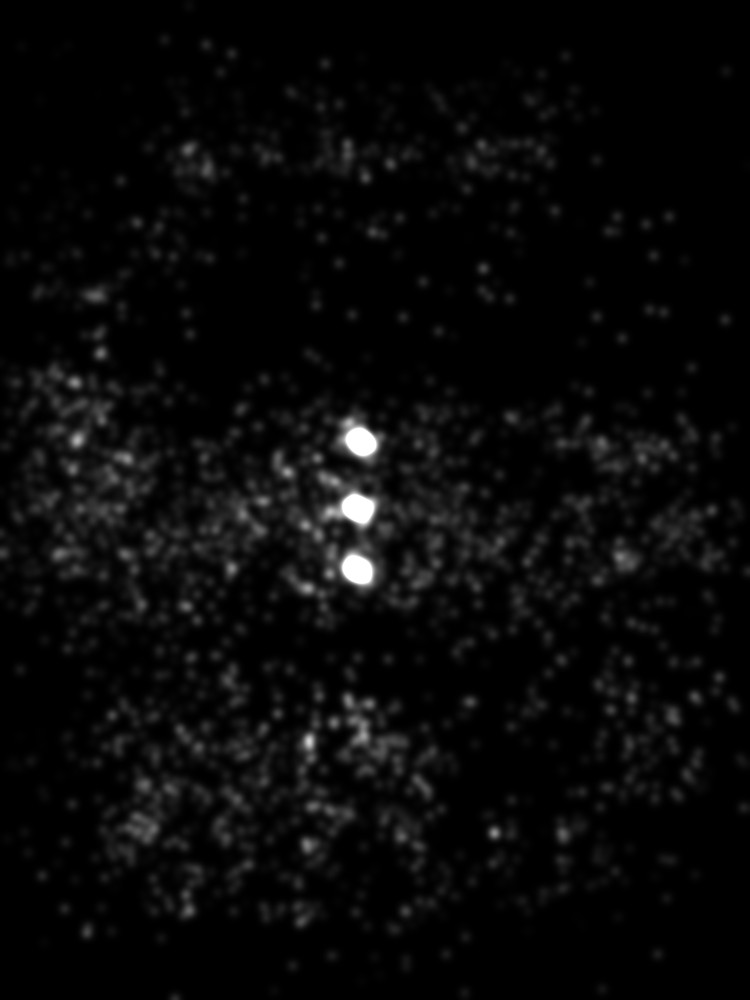}}
  \centerline{(b) Blurred and normalized vote image}\medskip
  \label{fig:votemap_obj2}
\end{minipage}
\hfill
\begin{minipage}[!htb]{.3\linewidth}
  \centering
  \centerline{\includegraphics[width=5cm]{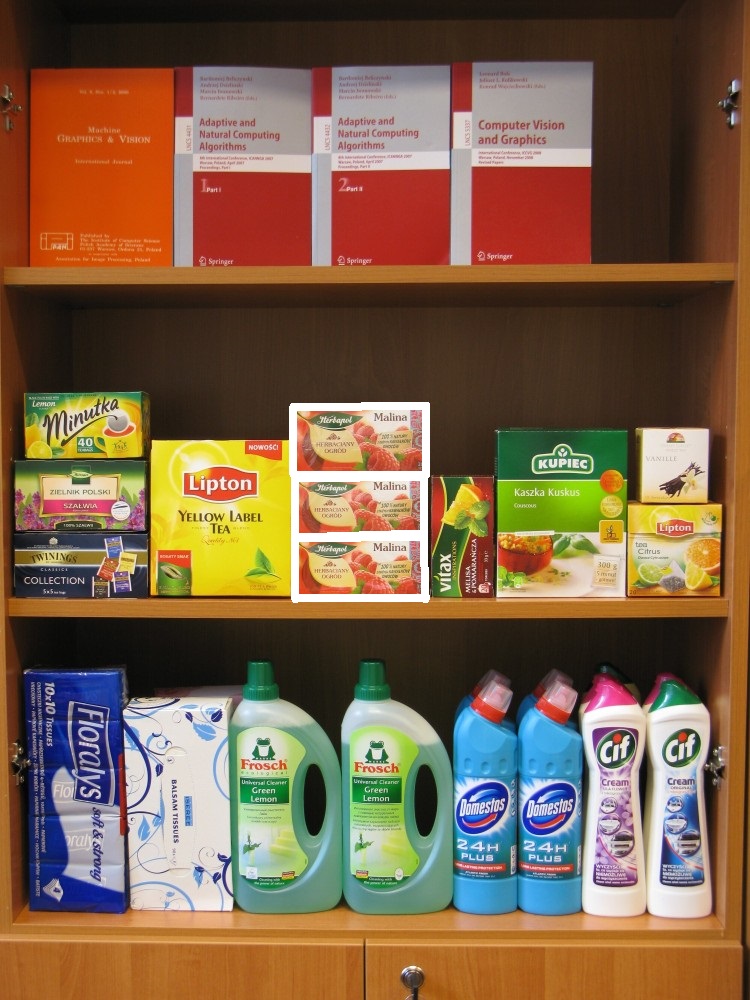}}
  \centerline{(c) Scene with tea detections}\medskip
  \label{fig:votemap_obj3}
\end{minipage}
\caption{Sample of vote image generated while localizing red, herbal tea casing.}
\label{fig:votemap}
\end{figure*}

Adjacency values are gathered in a single channel, gray vote image. Vote image can be optionally normalized for visualization purposes. Such normalized image has been shown in Figure~\ref{fig:votemap}. If the feature extraction and matching process are highly discriminative, the object instances in the scene should be recognizable by a human. Manual verification of vote image gives some level of valuable insight into votes intersperse in the image and a level of a false votes groupings recognizable by a human. 

Last step of Algorithm \ref{alg:prop_creation} is the search for propositions in a vote image. Proposition is a point in the vote image and corresponding part of vote space, where the potential object's center is located. We used Good Features To Track by Tomasi and Shi~\cite{Goodfeatures01} to detect multiple local maximas in the vote image and used them as the propositions. The number of propositions should be much higher than a number of objects in the image. It is trivial to set the Good Features To Track to find all the important points in the image, but it leads to generation of thousands of propositions. Number of propositions will significantly impact the algorithm's processing time, so it is not possible to ignore the need for a trade-off in detector's parameter adjustment. For each proposition's X and Y position the adjacency sum for corresponding votes in vote space is calculated and used for sorting purposes in line \ref{line:prop_sorting}. The highest adjacency value proposition should be the first taken later into a vote aggregation process. As the adjacency sum is proportional to the channel value in the vote image, the cue for sorting stage is easy to compute. Sorting the propositions ensures, that the strongest vote grouping will be processed first. In case of the positive object recognition, the vote data corresponding to object's detection area will be erased from vote space and image.

\subsection{Vote aggregation}

\noindent Second part of our approach contains a vote aggregation mechanism. Vote aggregation starts from a proposition's position, which should be the center of a local vote grouping in the vote image. Data of the vote groupings can significantly vary for different object instances in the scene. The best instances can be represented by hundreds of votes, when the weakest positive object response can be connected with only a few. Generic clustering may ignore such clusters and merge it with the bigger ones. Generic clustering algorithms has generic parameters, that are hard to adjust with object-oriented logic or even intuition. Some clustering approaches tend to cluster all the available vote data, even if the noise (false correspondences) fills most of the vote space.

We propose an iterative 2-pass vote aggregation process for selective clustering purposes. In each pass the unique filtering and the cascade filtering take place, which reject false positive detections. Two pass design prevents situations in which aggregation area contains multiple objects. Pass one of the aggregation collects all the votes in local area of proposition's position (line \ref{line:pass1_1} of Alg. \ref{alg:aggregation}). The size of a local area may be a function of a corresponding pattern size. After gathering of all the votes in the local area, the unique filtering is performed and the resulting group of votes is tested with a cascade of filters (lines \ref{line:pass1_2} and \ref{line:pass1_3} of Alg. \ref{alg:aggregation}).

In the second pass of the process the aggregation is conducted with Flood Fill algorithm, starting from the proposition's position (line \ref{line:pass2_1} of Alg. \ref{alg:aggregation}). The Flood Fill range is limited to a scaled down object's area. Such limitation can be constructed with a scale and rotation estimation from the first pass of the aggregation. The limitation ensures, that the aggregation process will not collect the votes from neighbouring object instances. Second pass of the algorithm contains unique filtering and cascade filtering as well, as the vote collection may be different in this pass.

For each group of votes, a unique filtering should be performed in each pass (lines \ref{line:pass1_2} and \ref{line:pass2_2} of Alg. \ref{alg:aggregation}). Unique filtering preserves only one vote with the highest adjacency corresponding to the same feature point in the pattern. We can do so, because we want the aggregated votes to be connected with only one object. If multiple votes are connected with one specific feature in the pattern we can assume that only the strongest vote isn't the noise.

Most of the feature point detectors incorporate mechanisms of rejecting the points located along the edges. Unfortunately, this mechanisms work only in a micro scale. In high resolution some graphical structures, that for human seem as a straight edge, has a very complicated, uneven shape for characteristic points detector. Characteristic points located along the edges have similar features, so may be matched with the same feature in the pattern. It leads to generation of many false propositions, which can sometimes be accepted by a cascade filters. 

Some of the false positive detections in our experiments were initiated as a bunch of feature points placed along a simple, steep gradients and edges. For instance, when the scene presented product shelves, more than a half of false detections contained edge of the shelf near the center and its vote data present mostly along the shelf's edge.

\subsection{Cascade filtering}

\noindent Cascade filtering (lines \ref{line:pass1_3} and \ref{line:pass2_3} of Alg. \ref{alg:aggregation}) is a process of validating vote group with a cascade of filters. Each filter can accept aggregated votes or reject them. Any rejection will result in dropping the aggregation process and removing the processed proposition from the propositions sorted queue. No vote data is removed from vote space or vote image in that situation. If all the filters in first pass accepts the vote group, process may estimate the size and rotation of the object represented by the majority of votes (line \ref{line:pass2_4} of Alg. \ref{alg:aggregation}). 

Cascade filters comprise of: (1)~vote count thresholding, (2)~adjacency sum thresholding, (3)~scale variance thresholding, (4)~rotation variance thresholding, (5)~feature points binary test, (6)~global normalised luminance cross correlation thresholding.
First pass of the vote aggregation uses filters: (1), (2), (3) and (4). Second pass of the process uses filters: (3), (4), (5) and (6).

Vote count thresholding is a simple filter, thresholding number of votes in the aggregated group. Lowe in his work~\cite{Sift02} proposed generalised Hough transform for object detection. In this method he has assumed that only three votes are enough to identify the object. Unfortunately, such assumption leads to many false positive detections. Three local features are not enough to describe complex, generic graphics. We tested vote count thresholding for values from 3 up to 20. We found 6 as the optimal value for filtering out too weak responses. If the vote grouping represents real object instance and has less than 6 votes, it means that the prior algorithm processes has too low effectiveness.

Adjacency sum thresholding rejects all the groups of votes with sum of adjacency values less than a threshold value. This filter in certain circumstances can be used instead of the vote count thresholding. Nevertheless the rejection data from these two filters may give an insight into vote certainty levels of the detection. Even huge vote groupings with more than 100 votes may have a very low adjacency sum value.

Scale variance thresholding rejects all the groups of votes with a scale value variance higher than the threshold value. One may rebuild this filter into mechanism separating noise signal from positive detection signal with a Gaussian model. For our purposes such method is computationally too expensive. Simple variance thresholding rejects many false detections and is easy to compute.

Rotation variance thresholding rejects all the groups of votes with rotation value's variance higher than the threshold value. Rotation variance thresholding works analogically to scale variance thresholding but using the rotation values. A rotation variance is not straightforward to compute. We set twelve buckets for rotation values and choose the three buckets with the highest count number. Its resultant was taken as an average rotation. All the values has been rotated so the average rotation was assigned to 180 degrees. Then the variation in regards to 180 degrees has been computed and used for thresholding.

Feature binary test uses feature points correspondence data preserved in each vote. We created multiple luminance binary tests for random feature pairs in the scene, which are represented by votes in aggregated votes group. We created identical tests for corresponding feature points on the pattern side. Each set of binary tests provided a binary string that can be compared with a hamming distance. The normalised distance can be thresholded. 

Normalised luminance cross correlation is used as a last filter. It needs the exact object's graphics patch extracted from the scene. It's computationally expensive, but can filter out many false positive detections, that cannot be filtered by previous filters. The images are resized to the size of 50x50 pixels before the calculation of the cross correlation. The filtering is conducted only in the second pass of the aggregation process, where the theoretical object's frame can be calculated from the data from the first pass.

\section{\uppercase{Experiments}}

\begin{figure*}[!htb]
\begin{minipage}[!htb]{.3\linewidth}
  \centering
  \centerline{\includegraphics[width=2cm]{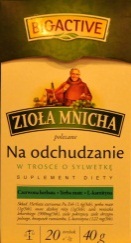}}
  \centerline{(a) Pattern}\medskip
  \label{fig:ziola_obj1}
\end{minipage}
\hfill
\begin{minipage}[!htb]{.6\linewidth}
  \centering
  \centerline{\includegraphics[width=10cm]{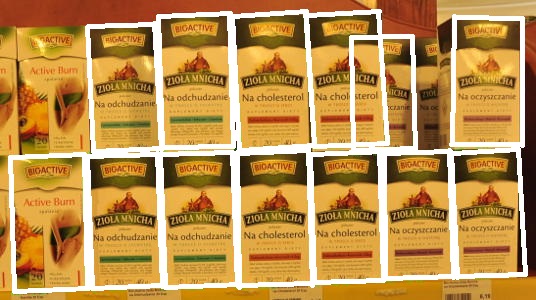}}
  \centerline{(b) Scene}\medskip
  \label{fig:ziola_obj2}
\end{minipage}
\caption{Sample of detection results}
\label{fig:ziola}
\end{figure*}

\noindent Our testing platform, incorporating method described in this paper, has been developed to search product logos and casings on scenes presenting market shelves and displays. The database used for the test for this paper consists of 120 shelf photos taken in 12MPx resolution and scaled down to 3MPx for testing purposes. The pattern group consists of 60 generic patterns of logos and product wrappings. Each shelf photo was tested with each one of the patterns, giving 7200 detection processes. The photos contained usually three classes of products so most of the patterns could generate only false positive detections. Average number of products presented in the scenes was \emph{23.6}. Patterns were scaled to have the bigger size between 512 and 256 pixels. In the application of product search on the market shelves we describe high quality images as photos bigger than 2MPx, with a minimum of ten thousands pixels for the smallest searched object and all of the logos text readable for a human. Our aggregation approach bases its effectiveness upon chosen local features. We used SIFT implementation for main experiments. Main advantage of our method lays in filtering out false detections and processing all possible occurrences. SIFT is a state-of-the-art features detector and descriptor. Our test showed that 100\% of the actual object instances were processed through our cascade filtering with a proper proposition's location. That's thanks to dense proposition detections and a straightforward vote image creation.

Detection effectiveness lays in proper vote group filtering. Amount of positive detections rejected during cascade filtering results from all the computer vision algorithms incorporated into detection system and can be hardly used to measure aggregation effectiveness without proper comparisons with similar methods in the same application field. False detection rate yields more analytical data. We found no false positive detections during our tests, that were fault of insufficient description capability of feature descriptor. All of false detections were the result of too loose parameters, that were needed for very high positive detection rate. Nevertheless we came across 203 false detections in 129 of 7200 detection processes, resulting in more than 1\%~(Table~\ref{tab:results}) false detection chance per detection process. This result seems low, but at the same time means \emph{66.2\%} chance, that the false detection will take place when looking for any product instance from our patterns database. 

Our method has been compared to the method using the HOG descriptor. For the training stage we generated set of 60 derivative images for each pattern through small affine transformations. We used all other patterns as a negative images. We used implementation of HOG method, called Classifier Tool For OpenCV and FANN \cite{HOGApp}. Our method achieved only slightly better detection rate, but significantly lower chance for false detections. The average number of false detections were almost two times higher for the HOG approach (Table~\ref{tab:results2}).

\begin{table}[h!]
\begin{center}
\begin{tabular}{|l|c|c|}
\hline
Method & Detection Rate & False Detection Chance\\
\hline
Ours & 81.3\% & 1.79\% \\
HOG & 73.6\% & 21.42\% \\
\hline
\end{tabular}
\end{center}
\caption{Detection rate and false detection chance for our tests.}
\label{tab:results}
\end{table}

\begin{table}[h!]
\begin{center}
\begin{tabular}{|l|c|}
\hline
Method & Average Number of False Detections\\
\hline
Ours  & 1.57\\
HOG  & 3.18\\
\hline
\end{tabular}
\end{center}
\caption{Average number of false detections for process, where the false detection occurred.}
\label{tab:results2}
\end{table}

During experiments with product casings we encountered number of problems with association of detections to a specific result group. Some products are very similar, with only slight local graphical differences. This is particularly true for the same brand with different aromas or casing sizes. Figure~\ref{fig:ziola} presents one of such cases, where tea casing has identical logo for its few variations with one being visually very different from the others. We decided to interpret only the visually off tea as a false detection. In retail field the rest of the detections should be processed further to discriminate different variations of the products. One can use partial patterns with a bag-of-words approach on top of our aggregation method to do so.

\section{\uppercase{Conclusions}}

\noindent In this paper the method of vote aggregation designed for use in multi-object multi-detection systems has been introduced. Aggregation process yields promising results in tests, leading to analysis of each potential object in the image. The unique filtering leaves out many false object occurrence propositions and the cascade filtering rejects most of the false positive detections, that is crucial for presented application. System built upon the aggregation method can achieve more than 80\% detection rate with the false detection chance below 2\%. It is still far from industrial standards, but there are many places for improvement as well.

Presented method is designed to analyze very high quality images. Images processed in tests were taken by a hand, resulting in high amount of blurred and skewed visual data. The method of image acquisition should be analyzed further. In future work we will incorporate estimates of the best parameters for presented method as well as solve simple parametrization dependencies. We are going to test the system with a two-phase approach, where second phase of the detection would use the patterns extracted directly from the scene. The pattern size has too much impact on the detection rate, as the feature points approach works the best, when the objects in the scene and in pattern images have the same size. We are going to evaluate resizing options for better detection results.

\section*{\uppercase{Acknowledgements}}

\noindent This work was co-financed by the European Union within the European Regional Development Fund.

\bibliographystyle{apalike}
{\small
\bibliography{lingaro}}

\end{document}